%% file: main.tex
\documentclass[runningheads]{llncs}

\usepackage{eccv}

\usepackage{eccvabbrv}

\usepackage{graphicx}
\usepackage{booktabs}
\usepackage{multirow}
\usepackage{pifont}
\newcommand{\cmark}{\ding{51}}
\newcommand{\xmark}{\ding{55}}
\usepackage{makecell}
\usepackage{xcolor}
\usepackage{tikz}
\usepackage{cite}
\usetikzlibrary{calc}
\usepackage{dsfont}

\definecolor{valid_unknown}{RGB}{237,180,6}
\definecolor{gt_unknown}{RGB}{252,101,25}
\definecolor{pred_unknown}{RGB}{203,18,18}
\definecolor{yellow}{RGB}{255,212,0}
\definecolor{grey}{RGB}{74,74,74}

\usepackage[accsupp]{axessibility}

\usepackage{hyperref}

\usepackage{orcidlink}

\begin{document}

\title{Hyp2Former: Hierarchy-Aware Hyperbolic Embeddings for Open-Set Panoptic Segmentation}

\titlerunning{Hyp2Former}

\author{Yao Lu\inst{1} \and
Rohit Mohan\inst{1} \and
Florian Drews\inst{2} \and
Yakov Miron\inst{2} \and
Abhinav Valada\inst{1}}

\authorrunning{Y.~Lu et al.}

\institute{Department of Computer Science, University of Freiburg, Germany \and
Bosch Research, Robert Bosch GmbH, Germany}

\maketitle

\begin{abstract} Recognizing unknown objects is crucial for safety-critical applications such as autonomous driving and robotics. Open-Set Panoptic Segmentation (OPS) aims to segment known \textit{thing} and \textit{stuff} classes while identifying valid unknown objects as separate instances. Prior OPS approaches largely treat known categories as a flat label set, ignoring the semantic hierarchy that provides valuable structural priors for distinguishing unknown objects from in-distribution classes. In this work, we propose Hyp2Former, an end-to-end framework for OPS that does not require explicit modeling of unknowns during training, and instead learns hierarchical semantic similarities continuously in hyperbolic space. By explicitly encoding hierarchical relationships among known categories, the model learns a structured embedding space that captures multiple levels of semantic abstraction. As a result, unknown objects that cannot be confidently classified as known categories still remain in close proximity to higher-level concepts (e.g., an unknown animal remains closer to “animal” or “object” than to unrelated concepts such as “electronics” or “stuff”) and can therefore be reliably detected, even if their fine-grained category was not represented during training. Empirical evaluations across multiple public datasets such as MS COCO, Cityscapes, and Lost\&Found demonstrate that Hyp2Former outperforms existing methods on OPS, achieving the best balance between unknown object discovery and in-distribution robustness. We make the code and models soon publicly available at \url{https://hyp2former.cs.uni-freiburg.de}.


\end{abstract}

\section{Introduction}

Panoptic segmentation (PS)~\cite{kirillov2019panoptic} has achieved remarkable performance in closed-world settings~\cite{mohan2022perceiving, mohan2026up, luz2026latent}, where the set of classes is predefined and identical during training and inference. Most existing approaches rely on flat classifiers \cite{vodisch2023codeps, cheng2022mask2former, kappeler2024few, vodisch2024good} that treat categories as independent, permutation-invariant and unstructured, ignoring the rich semantic relationships among them, such as the natural similarities between \textit{cat} and \textit{dog} or \textit{baseball} and \textit{baseball bat} \cite{sinha2024learning,hindel2024taxonomy,hindel2025dynamic}. While this is often sufficient in closed-world PS, it becomes a major limitation in Open-Set Panoptic Segmentation (OPS)~\cite{hwang2021eopsn, mohan2025open}, where valid objects from unseen categories can appear at test time and must be segmented and identified as unknown. In this setting, the model must generalize beyond fixed labels and acquire a transferable notion of objectness from known categories to unknowns~\cite{nayal2023rba}. 

However, existing OPS approaches typically detect unknowns through per-pixel uncertainty and clustering \cite{hwang2021eopsn, gasperini2023u3hs,schmidt2025prior2former}. They do not explicitly model unknowns at the instance level and are sensitive to distributional shifts, often struggling to distinguish between unknown objects (countable instances), unknown stuff categories, and out-of-distribution (OOD) regions. As a result, predictions for unknowns can be unstable and semantically inconsistent. To address this problem, we explicitly encode hierarchical relationships among known categories at the instance level to learn object representations at varying levels of semantic granularity (e.g., object$\rightarrow$ animal $\rightarrow$ dog). This promotes a broader notion of objectness, enabling novel objects to align with higher-level semantic concepts and to be reliably detected as unknown even when their fine-grained class is unavailable. Furthermore, this enables reliable recognition of unknown objects without the need for pseudo-labels~\cite{Xu2022DDOPS, yin2024revisiting}, auxiliary data for unknowns~\cite{rai2024mask2anomaly, nayal2023rba}, or prior assumptions\cite{hwang2021eopsn} about unseen categories.

A second limitation concerns the choice of embedding space used for hierarchy-aware semantic alignment. Previous OPS methods operate in Euclidean space \cite{hwang2021eopsn, Xu2022DDOPS, gasperini2023u3hs}, whose polynomial volume growth restricts their ability to faithfully represent tree-like structures without distortion \cite{ganea2018hyperbolic}. In contrast, hyperbolic space exhibits exponential volume growth, naturally accommodating the exponential increase in nodes with depth in hierarchical trees \cite{mettes2024hyperbolic}. Hyperbolic embeddings have been shown to yield compact and semantically coherent representations for structured domains such as trees and acyclic graphs \cite{sala2018representation, sarkar2011low}, and have since been successfully adapted to vision tasks involving hierarchical data \cite{nickel2017poincare,lang2022hyperbolic,khrulkov2020hyperbolic, hindel2025dynamic}.

To address these limitations, this work investigates a novel approach to OPS that learns hierarchical embeddings and class representations in hyperbolic space. Specifically, we extend Mask2Former~\cite{cheng2022mask2former} by projecting mask query embeddings into the hyperbolic Lorentz model, enabling continuous hierarchical representation learning. Classification and mask prediction remain in Euclidean space to preserve class discrimination and mask quality. We further introduce a novel hierarchical proxy-anchor loss that encourages query embeddings to align not only with class-specific proxies but also with their ancestral proxies in hyperbolic space. We exploit the intuition that embeddings of novel instances are more likely to lie in close proximity to some semantic node within the hierarchy, rather than collapsing toward background representations. This enables reliable recognition of unknown objects in dynamic environments without prior assumptions during training. 

Our primary contributions are summarized as follows: (1) We introduce Hyp2Former, a hierarchy-aware framework that learns semantically structured visual representations in hyperbolic space for OPS. (2) We propose a hierarchical proxy-anchor loss that aligns query embeddings with both class-specific and ancestral proxies, encouraging abstraction-aware representations and improved generalization to unknown instances. (3) We demonstrate that Hyp2Former achieves state-of-the-art performance for OPS on multiple benchmarks, with consistent performance improvement on both known categories and unknown objects. (4) We present extensive qualitative evaluations on different robot embodiments and sensing platforms. (5) Open-source code and trained models will be made available upon publication.

\section{Related Work}

{\parskip=0pt
\noindent\textbf{Close-World Panoptic Segmentation (PS)}:  
PS~\cite{kirillov2019panoptic} assigns each pixel in an image a semantic label for either \textit{stuff} classes (amorphous regions) or \textit{thing} classes (countable objects), and additionally predicts instance identities for \textit{thing} classes, thereby unifying semantic segmentation~\cite{hindel2025label, hurtado2026hyperspectral, mohan2026forecastocc} and instance segmentation in a single scene representation, within the broader family of dense prediction tasks~\cite{luz2024amodal}. Early work such as Panoptic-DeepLab \cite{cheng2020panoptic} followed a bottom-up approach that predicts semantic maps together with instance cues to group pixels into individual objects. More recent approaches shifted to mask-based, set-prediction formulations: MaskFormer~\cite{cheng2021maskformer} recasts PS as mask classification, predicting a set of segments with associated class labels, and Mask2Former~\cite{cheng2022mask2former} strengthens this paradigm by introducing masked attention. Under the closed-world setting, these methods achieve strong performance by learning discriminative representations for a fixed set of categories, but inherently assume that all test-time regions belong to the known label space.}

{\parskip=2pt
\noindent\textbf{Open-Vocabulary Methods (OV)}:  Recent OV approaches leverage pretrained vision–language models (VLM) to align visual and textual representations within a shared embedding space, enabling OV segmentation through textual prompts at inference time \cite{radford2021clip, carion2025sam3}. Despite their strong performance, they implicitly assume that the VLM encodes all categories encountered during inference, thereby restricting its recognition scope to the pretraining corpus \cite{gasperini2023u3hs}. Furthermore, it relies on textual prompts or visual cues which may not be fully available for unknowns at inference time. Prompt-free variants similarly depend on additional language models to generate candidate labels, maintaining reliance on pretrained supervision rather than achieving fully open-set recognition \cite{conti2026vocabulary, reichard2025open}.}

{\parskip=2pt
\noindent\textbf{Open-set Panoptic Segmentation (OPS)}: 
OPS extends panoptic segmentation by jointly segmenting known \textit{things} and \textit{stuff} while additionally identifying unknown objects as instances. Existing methods can be categorized based on whether OOD or Synthetic~\cite{mohan2024syn} data are exposed during training, thereby introducing distributional priors over unknowns (e.g., through extra datasets or void (unlabeled) regions). This corresponds to detecting “known unknowns” as termed by Bendale and Boult \cite{bendale2016towards}, rather than entirely unseen categories. Early approaches such as EOPSN \cite{hwang2021eopsn} and DDOPS \cite{Xu2022DDOPS} expose the model to void data during training and explicitly learn an unknown category. M2A \cite{rai2024mask2anomaly}, PoDS \cite{mohan2024panoptic}, and RbA \cite{nayal2023rba} similarly rely on OOD supervision. M2A and PoDS learn an additional background/unknown mask, whereas RbA performs mask rejection after tuning on such data. Recently, U3HS \cite{gasperini2023u3hs} proposes a holistic segmentation setting that avoids using any OOD data to prevent leakage of knowledge about unseen categories. U3HS employs a Dirichlet prior to cluster uncertain embeddings, and P2F \cite{schmidt2025prior2former} further models prediction uncertainty via a beta prior. Both U3HS and P2F perform anomaly segmentation using uncertainty measures and subsequently cluster the resulting regions using DBSCAN \cite{ester1996density} to obtain object instances. Our work also addresses this more challenging and generalizable setting, where no prior knowledge or auxiliary data for unseen categories are available during training. In contrast to prior approaches, we perform mask-based unknown detection directly at the instance level, eliminating the need for a separate clustering stage.}

{\parskip=2pt
\noindent\textbf{Hyperbolic Deep Learning in Computer Vision}: Hyperbolic representations have been explored to encode hierarchical semantic structure in visual data. Khrulkov et al.~\cite{khrulkov2020hyperbolic} introduced hyperbolic image embeddings for image classification and few-shot learning, demonstrating improved semantic organization of class relationships. Subsequent works extended hyperbolic embeddings to object detection~\cite{ge2023hyperbolic} for structuring object-level representations, and to open-set object detection in Hyp-OW~\cite{doan2024hypow}, where hierarchical hyperbolic embeddings are leveraged to generate pseudo-labels for unknown classification. For dense prediction, Atigh et al.~\cite{atigh2022hyperbolic} investigated hyperbolic semantic segmentation, demonstrating improved efficiency by operating per-pixel classification in hyperbolic space. This work focuses on the comparatively less explored setting of OPS, and extends hyperbolic representations to dense open-world scene understanding.}

\section{Method}
\subsection{Preliminaries}

{\parskip=0pt
\noindent\textbf{Mask Transformer for OPS}: 
We adopt Mask2Former \cite{cheng2022mask2former} as the base architecture for open-set panoptic segmentation due to its performance and end-to-end formulation. Multi-scale features extracted from the backbone are processed by a pixel decoder and a transformer decoder. A set of learnable queries is fed into the transformer decoder, which yields query embeddings. Each query predicts a class probability distribution and an associated mask embedding, trained via bipartite matching. During inference, Mask2Former applies mask-based post-processing by retaining queries whose confidence (maximum known-class probability) exceeds a predefined threshold. The remaining query masks compete in a pixel-wise manner, where overlapping regions are resolved based on confidence.}

{\parskip=2pt
\noindent\textbf{Hyperbolic Representations}: 
Hyperbolic space is a negatively curved non-Euclidean geometry exhibiting exponential volume growth, in contrast to the polynomial growth of Euclidean space~\cite{mettes2024hyperbolic}. This enables compact and low-distortion representations of hierarchically structured data, whose size increases exponentially with depth. Among several isometric models of hyperbolic space \cite{cannon1997hyperbolic}, we adopt the Lorentz model following~\cite{desai2023meru}. It represents $n$-dimensional hyperbolic space as the upper sheet of a two-sheeted hyperboloid embedded in $\mathbb{R}^{n+1}$. Each point $\mathbf{x} \in \mathbb{R}^{n+1}$ is decomposed into spatial and temporal components, written as $\mathbf{x} = [\mathbf{x}_{\text{space}}, x_{\text{time}}]$, where $\mathbf{x}_{\text{space}} \in \mathbb{R}^{n}$ and $x_{\text{time}} \in \mathbb{R}$. The geometry is defined through the Lorentzian inner product, 
\begin{equation}
\langle \mathbf{x}, \mathbf{y} \rangle_{\mathbb{L}}
=
- x_{\text{time}} y_{\text{time}}
+
\langle \mathbf{x}_{\text{space}}, \mathbf{y}_{\text{space}} \rangle_{\mathbb{E}}.
\label{eq:inner}
\end{equation}
which assigns a negative contribution to the time component and applies the Euclidean inner product $\langle \cdot , \cdot \rangle_{\mathbb{E}}$ to the spatial components. For a constant curvature $-c$ with $c > 0$, the Lorentz model is defined as
\begin{equation}
\mathbb{L}^{n}
=
\left\{
\mathbf{x} \in \mathbb{R}^{n+1}
\;:\;
\langle \mathbf{x}, \mathbf{x} \rangle_{\mathbb{L}} = -\frac{1}{c}
\right\}.
\label{eq:lorentz}
\end{equation}
The geodesic distance between two points $\mathbf{x},\mathbf{y} \in \mathbb{L}^n$, corresponding to the shortest path on the manifold, is given by:
\begin{equation}
d_{{\mathbb{L}}}(\mathbf{x}, \mathbf{y})
=
\frac{1}{\sqrt{c}}
\cosh^{-1}\!\left(-c \, \langle \mathbf{x}, \mathbf{y} \rangle_{\mathbb{L}} \right).
\label{eq:distance}
\end{equation}

To project a query embedding vector $\mathbf{v}$ from Euclidean space to the hyperbolic manifold, we apply the exponential map at the tangent space of the hyperboloid origin $\mathbf{O}$, following~\cite{desai2023meru}:
\begin{equation}
\mathbf{x}_{\text{space}}
=
\frac{\sinh\!\left(\sqrt{c}\,\|\mathbf{v}\|\right)}
{\sqrt{c}\,\|\mathbf{v}\|}
\mathbf{v}.
\end{equation}
The corresponding $x_{\text{time}}$ is then computed as $x_{\text{time}} = \sqrt{\frac{1}{c} + \|\mathbf{x}_{\text{space}}\|^2}$ according to Eq. \ref{eq:inner} and \ref{eq:lorentz}. For computing the average of points on the Lorentz model, we use the closed-form solution proposed by Law et al~\cite{law2019lorentzian}.

}

\subsection{Hyp2Former}

\begin{figure}[htb]
  \centering
  \includegraphics[height=3.8cm]{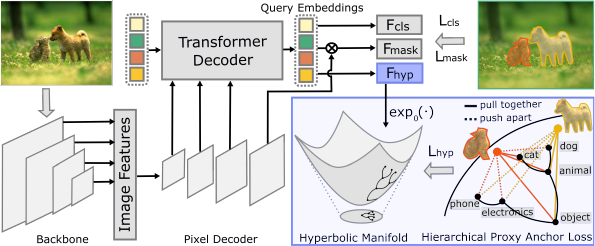}
  \caption{\textbf{Overview of the proposed Hyp2Former.} Multi-scale features are extracted from a backbone and are processed by a pixel and transformer decoder. A set of learnable queries interacts with the decoder to produce query embeddings, which are fed into the classification head $F_{cls}$ and mask head $F_{mask}$. In parallel, the embeddings are projected into a hyperbolic manifold, where an explicit semantic hierarchy guides the hierarchical proxy anchor loss $L_{hyp}$. This loss pulls embeddings toward their positive proxies (corresponding to the ground-truth class and its ancestors at different levels of granularity) and pushes them away from all other proxies, thereby structuring the hyperbolic embedding space according to semantic similarity. The predefined hierarchy $H_L$ with $L$ levels comprises three subtrees: \textit{object} (thing), \textit{background} (stuff), and an \textit{ignored} class for invalid regions. For clarity, only the \textit{object} subtree is illustrated.}
  \label{fig:Hyp2Former}
\end{figure}
We propose Hyp2Former, which adapts Mask2Former~\cite{cheng2022mask2former} to the open-world setting by leveraging hyperbolic hierarchical representation learning. Mask2Former inherently predicts class-agnostic mask proposals via learnable queries, which Hyp2Former directly exploits~\cite{cheng2022mask2former, ungur2025openmamba}. Hyp2Former jointly optimizes objectives in complementary geometric spaces. Standard classification and mask losses are optimized in Euclidean space to preserve discriminative separation among known classes and maintain strong in-distribution performance, while a hierarchical objective in hyperbolic space encodes continuous semantic relationships induced by a predefined hierarchy. This encourages embeddings to capture higher-level abstractions beyond fine-grained categories, resulting in semantically structured representations across levels of abstraction. 
Unlike approaches which model unknowns as a separate semantic class~\cite{doan2024hypow, Xu2022DDOPS,yin2024revisiting}, our method does not enforce artificial similarity among unseen objects. Instead, hierarchical modeling in hyperbolic space provides structured priors by embedding instances according to their semantic abstraction. During inference, instances that are not confidently assigned to any known leaf class in Euclidean space but remain close to higher-level proxies are treated as unknown, allowing valid unseen objects to emerge naturally from the structured embedding space. Figure~\ref{fig:Hyp2Former} illustrates the proposed framework.

{\parskip=2pt
\noindent\textbf{Modeling Continuous Hierarchy}: 
We leverage a predefined semantic hierarchy to impose structured constraints on the learned embeddings, encouraging them to reflect semantic proximity across abstraction levels and to encode a generalized notion of objectness that supports the identification of unknown objects. To faithfully model hierarchical structure, similarity $s(\cdot,\cdot)$ is defined in hyperbolic space using the negative Lorentzian distance $-d_{\mathbb{L}}(\cdot,\cdot)$ given in Eq.~\ref{eq:distance}, yielding a continuous contrastive formulation. A set of learnable proxy embeddings is introduced in hyperbolic space, assigning each ground-truth category $c \in \mathcal{C}$ a learnable semantic anchor. Intuitively, for an embedding $\mathbf{x}_i$ belonging to class $y_i$, it is encouraged that the similarity $s(\cdot,\cdot)$ between $\mathbf{x}_i$ and its ground-truth proxy embedding $\mathbf{p}_{y_i}$ is higher than the similarity to unrelated proxies:
\begin{equation}
s(\mathbf{x}_i, \mathbf{p}_{y_i})
\;>\;
s(\mathbf{x}_i, \mathbf{p}_k),
\quad
\forall\, k \in \mathcal{C}, \; k \neq y_i .
\end{equation}
In order to incorporate hierarchical structure, the alignment target is extended beyond the ground-truth class to include its ancestors in the hierarchy. Given the hierarchy $H_L$ with $L$ levels, the proxy set is defined as 
$\mathcal{P} = \{\mathcal{P}_0, \ldots, \mathcal{P}_{L-1}\}$,
where $\mathcal{P}_\ell$ denotes the set of proxies at hierarchy level $\ell$, and $\mathcal{P}_0$ corresponds to the leaf proxies representing the known categories. Higher-level proxies 
$p^{\ell} \in \mathcal{P}_{\ell}$ with $\ell \in \{1,\ldots,L-1\}$  correspond to ancestor proxies and are dynamically constructed as the hyperbolic average~\cite{law2019lorentzian} of their child proxies in $\mathcal{P}_{\ell-1}$. They are therefore not optimized directly but recomputed on-the-fly during training. For a labeled embedding $(\mathbf{x}_i, y_i)$, we define the positive proxy set (for attraction) and the negative proxy set (for repulsion) across hierarchy levels as:
\begin{equation}
\mathcal{P}^{+}(y_i)
=
\left\{
p_{\pi_\ell(y_i)}^{\ell}
\right\}_{\ell=0}^{L-1},
\qquad
\mathcal{P}^{-}(y_i)
=
\mathcal{P}
\setminus
\mathcal{P}^{+}(y_i),
\end{equation}
where $\pi_\ell(y_i)$ denotes the ancestor of class $y_i$ at level $\ell$.  For $\ell=0$, the ancestor corresponds to the leaf proxy of the ground-truth class, i.e., $p_{\pi_0(y_i)}^{0} = p_{y_i}$. Thus, each embedding is encouraged to align not only with its leaf proxy but also with its ancestor proxies, while being separated from unrelated proxies. The hierarchical proxy-anchor loss for a single sample is defined as:
\begin{equation}
\begin{aligned}
\mathcal{L}_{\text{hyp}}(\mathbf{x}_i,y_i)
=
&\underbrace{
\log\!\Bigg(
1+
\sum_{\ell=0}^{L-1}
\exp\!\Big(
-\alpha_{\ell}
\big(
s(\mathbf{x}_i,p_{\pi_\ell(y_i)}^{\ell})
-\delta
\big)
\Big)
\Bigg)
}_{\text{pull toward leaf and ancestor proxies}}
\\[6pt]
&+
\underbrace{
\log\!\Bigg(
1+
\sum_{p \in \mathcal{P}^{-}(y_i)}
\exp\!\Big(
\alpha_{\ell(p)}
\big(
s(\mathbf{x}_i,p)
+\delta
\big)
\Big)
\Bigg)
}_{\text{push away from unrelated proxies}},
\end{aligned}
\end{equation}
where $\delta>0$ is a margin. We set the scaling factor $\alpha>0$ to be level-dependent, with $\alpha_0 = \alpha_{\text{leaf}}$ for leaf proxies and $\alpha_\ell = \alpha_{\text{anc}}$ for all ancestor levels $\ell \ge 1$, where $\alpha_{\text{anc}} < \alpha_{\text{leaf}}$. This enforces strong alignment at the leaf level while allowing softer constraints at coarser semantic levels.}

{\parskip=2pt
\noindent
\textbf{Overall Loss}: 
The final training objective is 
\begin{equation}
\mathcal{L} = \mathcal{L}_{\text{cls}} + \mathcal{L}_{\text{mask}} + \lambda\mathcal{L}_{\text{hyp}},
\end{equation}
where $\lambda > 0$ is a weighting hyperparameter. $\mathcal{L}_{\text{hyp}}$ operates in hyperbolic space to learn hierarchically structured embeddings. The optimization objectives are thus decoupled, with $\mathcal{L}_{\text{cls}}$ and $\mathcal{L}_{\text{mask}}$ following \cite{cheng2022mask2former} in Euclidean space.}

{\parskip=2pt
\noindent
\textbf{Hierarchy-Guided Unknown Mining}: In the Mask2Former~\cite{cheng2022mask2former} pipeline, only queries with high classification confidence for known categories participate in the mask competition. In contrast, we consider the remaining low-confidence queries that cannot be confidently assigned to any known class as potential unknown candidates. These queries are then filtered using the learned hierarchical proxies. For each embedding $\mathbf{e}$, we compute its hyperbolic distance to all proxies $\{\mathbf{p}_j\}$ and identify the nearest proxy
$\mathbf{p}^*(\mathbf{e}) = \arg\min_{j} d_{\mathbb{L}}(\mathbf{e}, \mathbf{p}_j)$. Unknown candidates are then selected using the following steps: 

\noindent 1) \textit{Object consistency. } If the nearest proxy $\mathbf{p}^*(\mathbf{e})$ belongs to the object subtree of the hierarchy, the candidate is considered object-consistent. 

\noindent 2) \textit{Hierarchy-level scoring. } For object-consistent candidates, we assign a hierarchy score $S_{\text{hier}}(\mathbf{e})$ based on the hierarchy level of the nearest proxy. Higher-level object ancestors receive larger scores than fine-grained leaf proxies. 

\noindent 3) \textit{Semantic divergence. } To estimate the confidence of this semantic alignment, we compute a divergence score $S_{\text{div}}(\mathbf{e})$ based on margin distances between the closest and second-closest proxies at all hierarchy levels. Larger margins indicate stronger separation from competing semantic concepts. The final objectness score used for semantic-based unknown mining is defined as:
\begin{equation}
s(\mathbf{e}) = S_{\text{div}}(\mathbf{e}) + \beta S_{\text{hier}}(\mathbf{e}),
\label{eq:objectness_score}
\end{equation}
where $\beta$ is a hyperparameter. Finally, the top-$K$ candidates with the highest $s(\mathbf{e})$ are selected as unknown instances and included into the mask competition. More details on the objectness score are provided in the supplementary material.}

\section{Experiments}
In the following experiments, we show that Hyp2Former improves the trade-off between unknown detection and known-category performance over prior methods, without requiring additional data on unknown classes.

\subsection{Setup}

{\parskip=0pt
\noindent
\textbf{Datasets and Metrics}: 
We evaluate Hyp2Former on three public datasets: Cityscapes\cite{cordts2016cityscapes}, Lost\&Found\cite{pinggera2016lost}, and MS COCO\cite{lin2014coco}, focusing on OPS without making prior assumptions or using auxiliary OOD data for unknowns. For evaluation metrics, we report panoptic quality (PQ), recognition quality (RQ), and segmentation quality (SQ) for known classes and the unknown class.

\textit{Cityscapes} \cite{cordts2016cityscapes} is a widely used urban scene understanding benchmark covering approximately 50 cities, with dense annotations for 8 thing and 11 stuff classes. We follow the standard split of 2,975 training and 500 validation images and treat all categories as known. 

\textit{Lost\&Found} \cite{pinggera2016lost} contains real-world driving scenes with physically placed OOD road hazards,  disjoint from Cityscapes training classes, and is commonly used with Cityscapes to evaluate small-object unknown detection~\cite{gasperini2023u3hs}. Further, due to the controlled placement of OOD objects, some scenes differ from those in Cityscapes, resulting in a distributional shift and making the task more challenging. Models are trained on Cityscapes and directly evaluated on Lost\&Found without fine-tuning, assessing their ability to detect unknown objects. 

\textit{MS COCO} \cite{lin2014coco} is a challenging large-scale dataset which comprises 80 \textit{thing} and 53 \textit{stuff} classes across diverse indoor and outdoor scenes. Following prior works \cite{hwang2021eopsn, gasperini2023u3hs,schmidt2025prior2former}, we designate the least frequent 20\% of \textit{thing} classes as unknown. Images containing these classes are removed from the training set, resulting in 98,112 training images. The original validation set is split into two subsets. The open-world validation set consists of 827 images that contain at least one unknown class and is used to evaluate unknown recognition. The remaining images form the closed-world validation set, which contains only known classes and is used to assess performance retention on known categories. 

Although no supervision is provided for the unknown classes during training on COCO, they remain compatible with the hierarchy defined over the known classes (e.g., bear as a descendant of animal). In contrast, the unknown objects in Lost\&Found lie largely outside the training taxonomy, enabling evaluation of generalization beyond the predefined hierarchy. For COCO, we follow the dataset’s superclass definitions, and for Cityscapes, we adhere to the taxonomy of \cite{hindel2024taxonomy}. The dataset hierarchies are provided in the supplementary material.}

{\parskip=2pt
\noindent
\textbf{Implementation Details}: 
We adopt Mask2Former \cite{cheng2022mask2former} as the base architecture. For all experiments, a ResNet50\cite{he2016ResNet} backbone is adopted to ensure fair comparison with prior methods \cite{gasperini2023u3hs,schmidt2025prior2former}. The number of queries is set to $Q = 100$ and the embedding dimension to 256. We use AdamW~\cite{Loshchilov2017DecoupledWD} with a global batch size of 96 and a learning rate of $6 \times 10^{-4}$, obtained via linear scaling from the original configuration (batch size 16, learning rate $1 \times 10^{-4}$) used in~\cite{cheng2022mask2former,schmidt2025prior2former}. For our method, we use the hyperbolic curvature $c = 0.1$. In the hierarchical proxy anchor loss, we use $\alpha_{leaf}=5, \alpha_{anc}=2.5$ and $\delta=0.5$ for margin. The loss weight is set to $\lambda=0.5$. For further implmentation details, please refer to the supplementary material.

}

\subsection{Quantitative Results}

\begin{figure}[h!]
  \centering
  \includegraphics[width=0.9\linewidth]{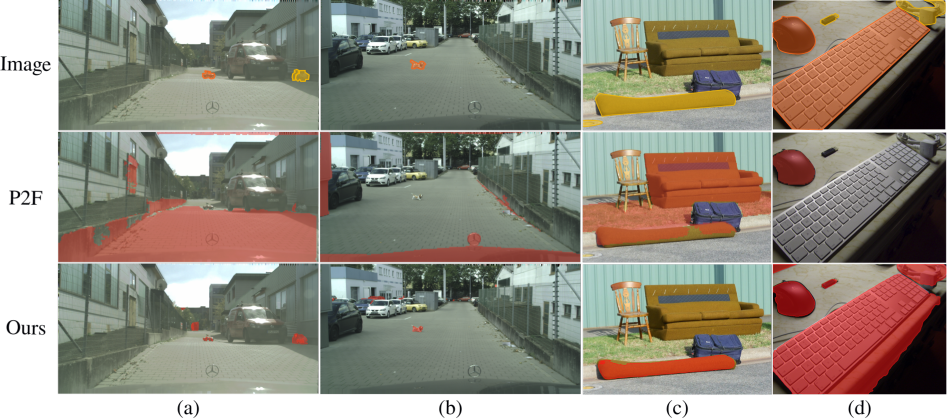}
\caption{\textbf{Qualitative comparison of unknown prediction.} The first two columns show results on Lost\&Found\cite{pinggera2016lost}, while the rest are on MS COCO \cite{lin2014coco}.  
Annotated unknowns are outlined in \textcolor{gt_unknown}{orange} and are reflected in the quantitative results, while additional valid but unannotated ones are outlined in \textcolor{valid_unknown}{yellow}. Predicted unknowns are shown in \textcolor{pred_unknown}{red}. Compared to P2F\cite{schmidt2025prior2former}, which often merges OOD \textit{stuff} regions and misses unknowns, our method yields more instance-consistent unknown predictions.}
  \label{fig:lf_qualitative}
\end{figure}

The performance on segmenting unseen unknown instances on Lost\&Found~\cite{pinggera2016lost} is reported in Table~\ref{tab:lf}. Our method achieves the highest PQ among all compared approaches, surpassing both earlier OPS methods and recent uncertainty-based frameworks that identify unknown objects without prior knowledge during training~\cite{gasperini2023u3hs, schmidt2025prior2former}. The improvement is primarily driven by higher RQ, indicating more accurate recognition and localization of unknown instances. These results highlight the benefit of hierarchical representation learning at the instance level for detecting objects outside the predefined taxonomy and training distribution. To further analyze the differences, Figure~\ref{fig:lf_qualitative}(a--b) presents a qualitative comparison with P2F~\cite{schmidt2025prior2former}. As an uncertainty-based method, P2F detects anomalous regions and subsequently clusters them into instances, which can lead to fragmented predictions and over-segmentation of OOD \textit{stuff} regions under distributional shifts. In contrast, our approach directly promotes instance-level consistency through structured hierarchical regularization, yielding more coherent predictions for unknown objects while suppressing irrelevant OOD regions. We additionally compare the run time and peak GPU memory with P2F\cite{schmidt2025prior2former}, measured on an NVIDIA GeForce RTX 4060 Ti with batch size 1. FPS is averaged over the Lost\&Found~\cite{pinggera2016lost}. Hyp2Former runs at 6.51 FPS with 6.21 GB peak memory, while P2F achieves 1.02 FPS with 7.40 GB. The slower speed of P2F is mainly due to its DBSCAN-based post-processing used to cluster anomaly regions into object instances.

\begin{table}[h!]
\centering
\caption{\textbf{Panoptic segmentation performance on unseen unknown instances.} All models are trained on Cityscapes~\cite{cordts2016cityscapes} and evaluated on Lost\&Found \cite{pinggera2016lost}. EOPSN~\cite{hwang2021eopsn} does not detect unknown instances in this cross-dataset setting, resulting in zero PQ. Hyp2Former achieves the highest performance. Best results are highlighted in \textbf{bold}. 
}
\setlength{\tabcolsep}{6pt}

\fontsize{8pt}{8pt}\selectfont
\begin{tabular}{l|l|ccc}
\toprule
\multirow{2}{*}{Model} & \multirow{2}{*}{Base Architecture} & \multicolumn{3}{c}{Unknown Classes} \\

 &  & PQ$\uparrow$ & RQ$\uparrow$ & SQ$\uparrow$ \\
\midrule
EOPSN \cite{hwang2021eopsn} & Panoptic FPN \cite{kirillov2019panopticFPN}
& 0.00 & 0.00 & 0.00 \\
OSIS \cite{wong2020identifying} & custom
& 1.45 & 2.23 & 65.11 \\
RbA \cite{nayal2023rba} & Mask2Former \cite{cheng2022mask2former}
& 7.30 & 10.17 & 71.85 \\
U3HS \cite{gasperini2023u3hs} & Panoptic-DeepLab \cite{cheng2020panoptic}
& 7.94 & 12.37 & 64.11 \\
EAM \cite{grcic2023advantages}  & Mask2Former \cite{cheng2022mask2former}
& 8.79 & 12.20 & 72.09 \\
M2A \cite{rai2024mask2anomaly}  & Mask2Former \cite{cheng2022mask2former}
& 9.91 & 13.49 & 73.45 \\
PoDS\cite{mohan2024panoptic}  & PAPS \cite{mohan2022perceiving}
& 10.96 & 16.09 & 68.13 \\
P2F \cite{schmidt2025prior2former}  & Mask2Former \cite{cheng2022mask2former}
& 11.22 & 15.06 & \textbf{74.47} \\
Hyp2Former (ours) & Mask2Former \cite{cheng2022mask2former}
& \textbf{12.15} & \textbf{17.11} & 70.97 \\
\bottomrule

\end{tabular}
\label{tab:lf}
\end{table}

To assess the impact of unknown recognition on in-distribution performance, we evaluate performance on known categories under both open-world and closed-world settings on Cityscapes (Table~\ref{tab:cityscapes}). In open-world settings, models predict both known and unknown, and in closed-set settings, only known classes are evaluated without detecting any unknowns, following \cite{gasperini2023u3hs, schmidt2025prior2former}. In the open-world setting, our method significantly outperforms prior work on known classes (> 8\% PQ improvement) while maintaining comparable closed-world performance. Importantly, it exhibits the smallest PQ drop ($\Delta$PQ) when transitioning from closed-world to open-world evaluation, indicating stable known-class performance under open-set conditions. Although P2F achieves slightly higher closed-world PQ, it suffers a significantly larger degradation in the open-world setting. This indicates that our approach more consistently preserves known-class predictions in the presence of unknown objects. Combined with the previously demonstrated improvements in unknown recognition on Lost\&Found (Table~\ref{tab:lf}), these results confirm that our method achieves the strongest overall performance for OPS without sacrificing in-distribution performance.

\begin{table}[t]
\centering
\caption{\textbf{Known-class performance in open-world setting on Cityscapes~\cite{cordts2016cityscapes}.}
Hyp2Former exhibits the smallest degradation and has strong known-class accuracy.}
\setlength{\tabcolsep}{4pt}
\fontsize{8pt}{8pt}\selectfont

\begin{tabular}{l|ccc|ccc|c}
\toprule
\multirow{2}{*}{Model} &
\multicolumn{3}{c|}{Open-World} &
\multicolumn{3}{c|}{Closed-World} &
\multirow{2}{*}{$\Delta$PQ$\downarrow$} \\

& PQ$\uparrow$ & RQ$\uparrow$ & SQ$\uparrow$ 
& PQ$\uparrow$ & RQ$\uparrow$ & SQ$\uparrow$ & \\

\midrule
U3HS \cite{gasperini2023u3hs} 
& 41.21 & 51.67 & 79.77 
& 46.53 & 58.99 & 78.87 
& 5.32 \\
P2F \cite{schmidt2025prior2former} 
& 45.25 & 54.68 & 80.69 
& \textbf{59.40} & \textbf{72.34} & 80.78 
& 14.15 \\

Hyp2Former (ours) 
& \textbf{53.91} & \textbf{65.20} & \textbf{81.38} 
& 59.20 & 71.96 & \textbf{80.97} 
& \textbf{5.29} \\

\bottomrule
\end{tabular}
\label{tab:cityscapes}
\end{table}

We further evaluate our method on COCO\cite{lin2014coco} to assess its robustness in more complex and diverse scenarios (Table~\ref{tab:coco}). Our model achieves the best performance on both unknown and known classes, evaluated on the open-world and closed-world validation sets, respectively. Notably, it establishes a clear margin over prior approaches on known classes while simultaneously improving unknown-class performance. Overall, these results demonstrate that the proposed hierarchy-aware hyperbolic representation learning provides a stable structural prior that remains effective across datasets with varying scene statistics.

\begin{table}[t]
\centering
\caption{
\textbf{Performance on unknown and known classes on COCO~\cite{lin2014coco}. }
Unknowns are evaluated with an open-world validation set and known classes on the closed-world validation set. 
Hyp2Former performs best in both settings.}
\setlength{\tabcolsep}{4pt}
\fontsize{8pt}{8pt}\selectfont

\begin{tabular}{l|ccc|ccc}
\toprule
\multirow{2}{*}{Method}  &
\multicolumn{3}{c|}{Unknown Classes} &
\multicolumn{3}{c}{Known Classes} \\

& PQ$\uparrow$ & RQ$\uparrow$ & SQ$\uparrow$ 
   & PQ$\uparrow$ & RQ$\uparrow$ & SQ$\uparrow$ \\
\midrule
EOPSN \cite{hwang2021eopsn}       & 0.01 & 0.01 & 74.55 & 36.50 & 45.06 & 76.32 \\
DDOPS \cite{Xu2022DDOPS}       & 8.39 & 10.29 & 81.52 & 37.93 & 46.80 & 77.28 \\
U3HS  \cite{gasperini2023u3hs}        & 9.62 & 13.20 & 72.84 & 22.03 & 27.92 & 69.03 \\
P2F \cite{schmidt2025prior2former}         & 10.71 & 13.39 & 79.96 
                         & 41.34 & 50.79 & 78.60 \\
Hyp2Former (ours)   & \textbf{12.39} & \textbf{15.04} & \textbf{82.39} 
                         & \textbf{47.20} & \textbf{56.86} & \textbf{81.00} \\
\bottomrule
\end{tabular}
\label{tab:coco}
\end{table}

\subsection{Qualitative Evaluations}
We present qualitative results illustrating Hyp2Former in real-world scenarios and the learned embedding structure, demonstrating robust unknown prediction under distributional shift and semantically organized hyperbolic embeddings. Additional qualitative results are provided in the supplementary material.

\begin{figure}[t]
  \centering
  \includegraphics[width = 0.9\linewidth]{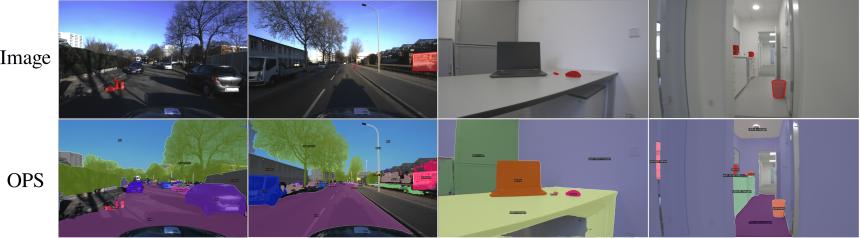}
  \caption{\textbf{Real-world zero-shot evaluation.} Unknown objects are marked in \textcolor{pred_unknown}{red}. Without fine-tuning, Hyp2Former generalizes to unseen real-world scenes and reliably detects unknown objects under distributional shift.
  }
  \label{fig:spot}
\end{figure}

{\parskip=2pt
\noindent
\textbf{Generalization to Real-World Scenarios}: 
Figure~\ref{fig:spot} shows that Hyp2Former generalizes in a zero-shot manner across different robot embodiments and sensing platforms beyond standard OPS benchmarks, without any further fine-tuning or adaptation.
Specifically, the COCO-trained model is directly applied to indoor data captured by a Boston Dynamics Spot quadruped robot, whereas the Cityscapes-trained model is evaluated on real-world urban driving scenes recorded with a vehicle-mounted FLIR Blackfly 23S3C RGB camera ($1920\times800$ resolution). Despite substantial shifts in viewpoint, sensor characteristics, and scene composition, the model consistently recognizes valid objects outside the training taxonomy and segments them as coherent unknown instances, while preserving stable panoptic predictions for known categories. This highlights the transferability of the learned hierarchical embeddings across real-world domains.}

\begin{figure}[t]
  \centering
  \includegraphics[width = 0.9\textwidth]{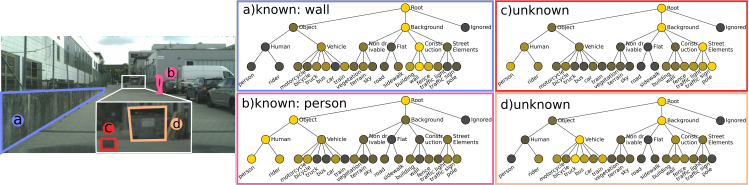}
  \caption{\textbf{Hierarchical distance visualization.}
Hyperbolic distances between mask embeddings and proxies are shown across hierarchy levels (closest in \textcolor{yellow}{yellow}, farthest in \textcolor{grey}{grey}). For unknown regions, the nearest leaf proxies remain interpretable, indicating that embeddings follow the learned hierarchy. 
In Cityscapes~\cite{cordts2016cityscapes}, \textit{pole} is defined as background (\textit{stuff}) rather than object (\textit{thing}), although it is countable.}
  \label{fig:tree_highlight}
\end{figure}

{\parskip=2pt
\noindent
\textbf{Hyperbolic Distance for Structured Reasoning}: 
To examine the learned embedding structure, we compute the hyperbolic distances between highlighted mask embeddings and all proxies in the hierarchy. At each hierarchical level, the closest proxy is highlighted in yellow and the most distant in grey, reflecting similarity through hyperbolic proximity. For known classes such as wall (Fig.\ref{fig:tree_highlight}a) and person (Fig.\ref{fig:tree_highlight}b), the embeddings consistently exhibit minimal distance to the correct leaf class and its corresponding ancestors across hierarchy levels. For unknown regions (Fig.\ref{fig:tree_highlight}c–d), although no exact leaf class is defined, the nearest leaf proxies remain interpretable and structurally consistent within the hierarchy. For example, the box (Fig.\ref{fig:tree_highlight}c) is closer to object-related proxy \textit{pole}, while the trash container (Fig.\ref{fig:tree_highlight}d) lies nearer to \textit{truck}, although at higher hierarchy levels it is slightly misaligned toward background (\textit{stuff}) rather than object (\textit{thing}). Importantly, these embeddings are not randomly distributed but remain structured relative to the learned hierarchy, suggesting that unknowns can be identified based on hyperbolic distance, as the embedding space preserves semantic structure.

}

\subsection{Ablation Studies}
In this section, we analyze the influence of different components, as well as the role of hierarchical supervision and hierarchy depth. Further ablation on hyperbolic curvature is provided in the supplementary material.

{\parskip=2pt
\noindent
\textbf{Influence of Different Components}: We analyze the contribution of each component on segmentation of unknown and known classes (Table ~\ref{tab:abl_unknown_mining}). Without hierarchical representation learning and unknown mining, the method reduces to Mask2Former~\cite{cheng2022mask2former}. The only difference lies in post-processing. Mask2Former retains only queries with high confidence in known classes for mask competition, whereas we apply the same known-class selection but assign all the remaining queries to the unknown class and include them in the competition. Under this setting, the model achieves the weakest unknown detection but the strongest in-distribution performance, as it focuses solely on known-category separation. This also shows that Mask2Former can generate valid object masks in a class-agnostic manner, even without explicit modeling of unknowns. Adding hierarchical embeddings improves unknown detection, with a slight decrease in known-class accuracy due to the additional structural constraint on the embeddings. Finally, semantic unknown mining ranks the filtered queries by the proposed objectness score (Eq. \ref{eq:objectness_score}) and selects the top-\textit{K} as unknown candidates, improving both unknown and known performance by reducing interference between them. Compared to the original Mask2Former baseline, the full model improves unknown detection by 26.17\%, with an 8.78 \% reduction in known-class performance.

}

\begin{table}[t]
\centering
\caption{\textbf{Ablation of the main components.} Models are trained on Cityscapes~\cite{cordts2016cityscapes} and evaluated on Lost\&Found~\cite{pinggera2016lost} (unknown classes) and Cityscapes (known classes) under the open-world setting. Combining hierarchical embeddings with  unknown mining achieves the best balance between unknown detection and known-class performance.}
\setlength{\tabcolsep}{4pt}
\fontsize{8pt}{8pt}\selectfont

\begin{tabular}{c|c|ccc|ccc}
\toprule
\multirow{2}{*}{\makecell{Hierarchical\\Regularization}}
 & 
\multirow{2}{*}{\makecell{Semantic\\Unknown Mining}} &
\multicolumn{3}{c|}{Unknown Classes} &
\multicolumn{3}{c}{Known Classes} \\

& & PQ$\uparrow$ & RQ$\uparrow$ & SQ$\uparrow$
   & PQ$\uparrow$ & RQ$\uparrow$ & SQ$\uparrow$ \\
\midrule
\xmark & \xmark 
& 9.63 & 14.23 & 67.67 & \textbf{59.10} & \textbf{71.44} & \textbf{81.54} \\
\cmark & \xmark

& \underline{11.33} & \underline{16.06} & \underline{70.55} & 53.84 & 65.13 & 81.34 \\
\cmark & \cmark 
& \textbf{12.15} & \textbf{17.11} & \textbf{70.97} & \underline{53.91} & \underline{65.20} & \underline{81.38} \\

\bottomrule
\end{tabular}
\label{tab:abl_unknown_mining}
\end{table}

{\parskip=2pt
\noindent
\textbf{Role of Hierarchical Structure}: 
We further asses whether the gains on unseen unknowns stem from stronger class-wise alignment or from explicit hierarchy via ancestor supervision. To this end, we compare a leaf-only variant, in which embeddings are aligned only with their ground-truth class proxy, to a full hierarchical variant, in which embeddings are additionally aligned with all ancestor proxies. As shown in Table \ref{tab:leaf}, regularizing only at the class level yields weaker unknown detection than the full hierarchical variant. While class-only supervision yields slightly better performance on known classes due to its milder constraint and closer alignment with the standard Euclidean class-separation objective, incorporating ancestor supervision substantially improves unseen unknown segmentation. This supports our motivation that hierarchical structure provides informative priors beneficial for OPS.}

\begin{table}[t]
\centering
\caption{\textbf{Impact of hierarchical structure.} Models are trained on Cityscapes~\cite{cordts2016cityscapes} and evaluated on Lost\&Found~\cite{pinggera2016lost} and Cityscapes. Adding ancestor proxy alignment improves unknown segmentation at a minor cost to known-class performance.}
\setlength{\tabcolsep}{4pt}
\fontsize{8pt}{8pt}\selectfont

\begin{tabular}{c|c|ccc|ccc}
\toprule
\multirow{2}{*}{$\mathcal{L}_{leaf}$} & \multirow{2}{*}{$\mathcal{L}_{ancestor}$} &
\multicolumn{3}{c|}{Unknown Classes} &
\multicolumn{3}{c}{Known Classes} \\

& & PQ$\uparrow$ & RQ$\uparrow$ & SQ$\uparrow$
   & PQ$\uparrow$ & RQ$\uparrow$ & SQ$\uparrow$ \\
\midrule

\cmark & \xmark & 7.83 & 11.35 & 69.03 & \textbf{55.67} & \textbf{67.65} & 80.88 \\
\cmark & \cmark & \textbf{11.33} & \textbf{16.06} & \textbf{70.55} & 53.84 & 65.13 & \textbf{81.34 }\\
\bottomrule
\end{tabular}
\label{tab:leaf}
\end{table}

{\parskip=2pt
\noindent
\textbf{Impact of Hierarchy Depth}: 
We further ablate the impact of taxonomy definition and taxonomy depth in Table~\ref{tab:hierarchy}. Here, H3, H4, and H5 denote hierarchies with three, four, and five levels, respectively, where H3 is the shallowest and H5 the deepest structure. In H3, all thing classes share a single high-level \emph{object} proxy. This coarse abstraction pulls diverse categories toward a common representation, weakening inter-class separation and reducing open-world performance on known classes. Increasing the depth to H4 introduces more meaningful intermediate abstractions, allowing semantically related classes to share structure without collapsing into a single global concept, and achieves the best overall trade-off. In contrast, H5 introduces an overly fine-grained hierarchy. Given the limited number of Cityscapes classes, some ancestors are supported by only a few children, weakening shared semantic priors and reducing unknown recognition. Overall, these results highlight the importance of selecting a balanced hierarchy depth that captures semantic structure without over-fragmentation.}

\begin{table}[t]
\centering
\caption{\textbf{Impact of hierarchy depth.} Models are trained on Cityscapes~\cite{cordts2016cityscapes} and evaluated on Lost\&Found~\cite{pinggera2016lost} and Cityscapes. A moderate depth (H4) performs best, while overly shallow (H3) or overly deep (H5) hierarchies reduce generalization.}
\setlength{\tabcolsep}{4pt}
\fontsize{8pt}{8pt}\selectfont

\begin{tabular}{c|ccc|ccc}
\toprule
\multirow{2}{*}{Hierarchy} &
\multicolumn{3}{c|}{Unseen Classes} &
\multicolumn{3}{c}{Open-World} \\

& PQ$\uparrow$ & RQ$\uparrow$ & SQ$\uparrow$
& PQ$\uparrow$ & RQ$\uparrow$ & SQ$\uparrow$ \\
\midrule

H3 & 8.00 & 11.78 & 67.88 & 52.23 & 63.48 & 80.92 \\

H4 & \textbf{11.33} & \textbf{16.06} & \textbf{70.55} &\textbf{ 53.84 }& 65.13 & \textbf{81.34 }\\

H5 & 7.66 & 11.02 & 69.51 & 53.76 & \textbf{65.34} & 80.75 \\
\bottomrule
\end{tabular}
\label{tab:hierarchy}
\end{table}

\section{Conclusion}
In this work, we introduced Hyp2Former, a hierarchy-aware framework for OPS that learns structured embeddings in hyperbolic space. By modeling semantic relationships between categories, the method enables reasoning about objectness beyond the fixed training taxonomy. Experiments show that Hyp2Former achieves state-of-the-art performance for OPS, improving unknown object discovery while maintaining strong performance on known classes. 
These findings underscore the potential of incorporating semantic relationships beyond flat classification, supporting a transferable notion of objectness that can help detect valid unknowns even when their categories were not observed during training.

{\parskip=2pt
\noindent
\textbf{Acknowledgements}: 
This research was funded by the German Research Foundation Emmy Noether Program grant number 468878300 and an academic grant from NVIDIA. Rohit Mohan was supported by the Bosch Research collaboration on AI-driven automated driving.}
\bibliographystyle{splncs04}
\bibliography{main}
\appendix
\renewcommand{\theHsection}{appendix.\arabic{section}}
\renewcommand{\theHsubsection}{appendix.\arabic{section}.\arabic{subsection}}
\renewcommand{\theHequation}{appendix.\arabic{equation}}

\input{supplementary}

\end{document}

%% file: supplementary.tex
\title{Supplementary Material \\[2mm]
Hyp2Former: Hierarchy-Aware Hyperbolic Embeddings for Open-Set Panoptic Segmentation}

\titlerunning{Supplementary Material}

\author{Yao Lu\inst{1} \and
Rohit Mohan\inst{1} \and
Florian Drews\inst{2} \and
Yakov Miron\inst{2} \and
Abhinav Valada\inst{1}}

\institute{Department of Computer Science, University of Freiburg, Germany \and
Bosch Research, Robert Bosch GmbH, Germany}

\maketitle

\section{Dataset Information}
In this section, we provide additional details including the hierarchy definitions used for each dataset. The hierarchy construction follows several general principles. First, the hierarchy contains a single root node whose immediate children correspond to objects (instance-level \textit{thing}), background (amorphous \textit{stuff}), and ignored regions (unmatched/invalid regions during training). Second, all leaf nodes correspond exactly to the ground-truth classes defined in the dataset. Third, the separation of leaf nodes into object and background categories follows the original \textit{thing}/\textit{stuff} distinction provided by the dataset annotations. Finally, we explicitly design intermediate parent nodes to group semantically related leaf-node classes into higher-level semantic concepts, ensuring a meaningful hierarchy between the top-level categories (object, background, ignored) and the leaf-level classes. Following the standard open-world evaluation protocol~\cite{hwang2021eopsn,gasperini2023u3hs,schmidt2025prior2former}, we assume that unknown objects belong to the thing category, while stuff categories remain closed-set.
\subsection{Cityscapes}
\begin{figure}[htbp]
\centering
\includegraphics[width=\linewidth]{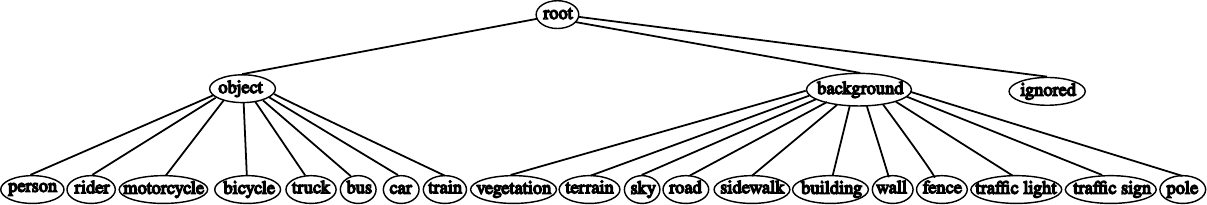}
\caption{Hierarchy definition for Cityscapes~\cite{cordts2016cityscapes} with 3 levels (H$_3$).}
\label{fig:H3}
\end{figure}

\begin{figure}[htbp]
\centering
\includegraphics[width=\linewidth]{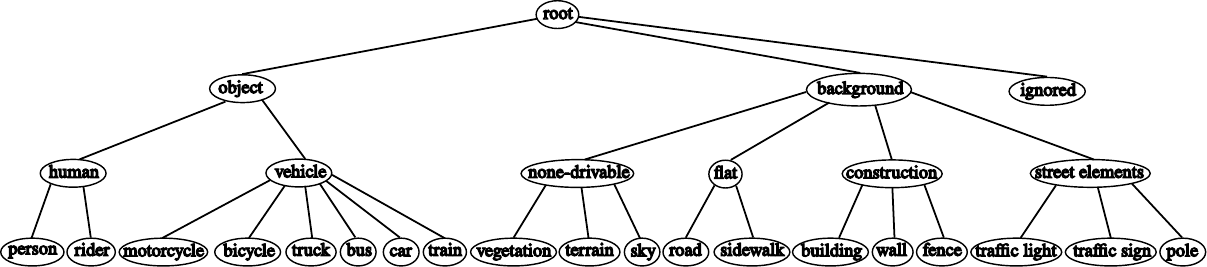}
\caption{Hierarchy definition for Cityscapes~\cite{cordts2016cityscapes} with 3 levels (H$_4$).}
\label{fig:H4}
\end{figure}

\begin{figure}[htbp]
\centering
\includegraphics[width=\linewidth]{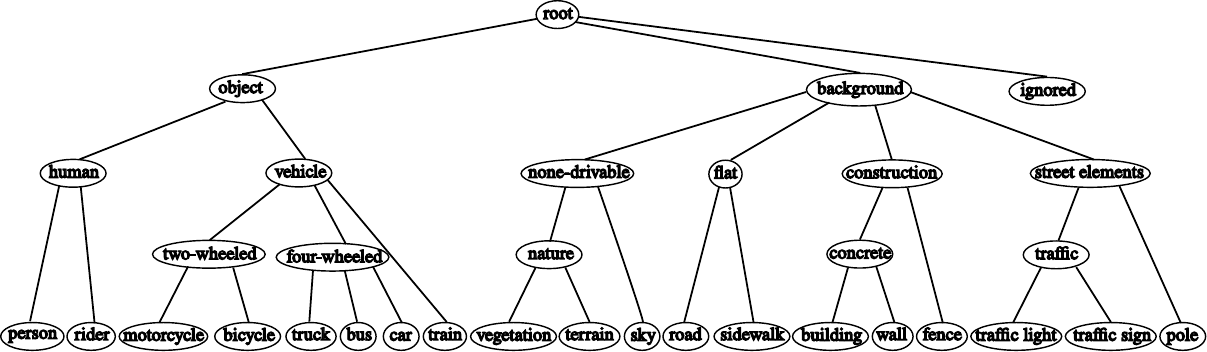}
\caption{Hierarchy definition for Cityscapes~\cite{cordts2016cityscapes} with 3 levels (H$_5$).}
\label{fig:H5}
\end{figure}

To ablate the influence of hierarchy definition and hierarchy depth, we train models using three hierarchies with different depths, namely H$_3$  (Figure~\ref{fig:H3}), H$_4$ (Figure~\ref{fig:H4}), and H$_5$ (Figure~\ref{fig:H5}). The hierarchy definition primarily follows those proposed by Hindel~et~al.~\cite{hindel2024taxonomy}, with slight modifications to align with the explicit separation of \emph{thing} and \emph{stuff} categories used in the original Cityscapes~\cite{cordts2016cityscapes} annotations. This adjustment is required since the original work by Hindel~et~al.~\cite{hindel2024taxonomy} focuses on semantic segmentation, which does not require distinguishing between \emph{thing} and \emph{stuff} categories to learn objectness. In contrast, this distinction is essential in our panoptic segmentation setting.

\subsection{COCO}

\begin{figure}[p]
\centering
\includegraphics[width=\linewidth]{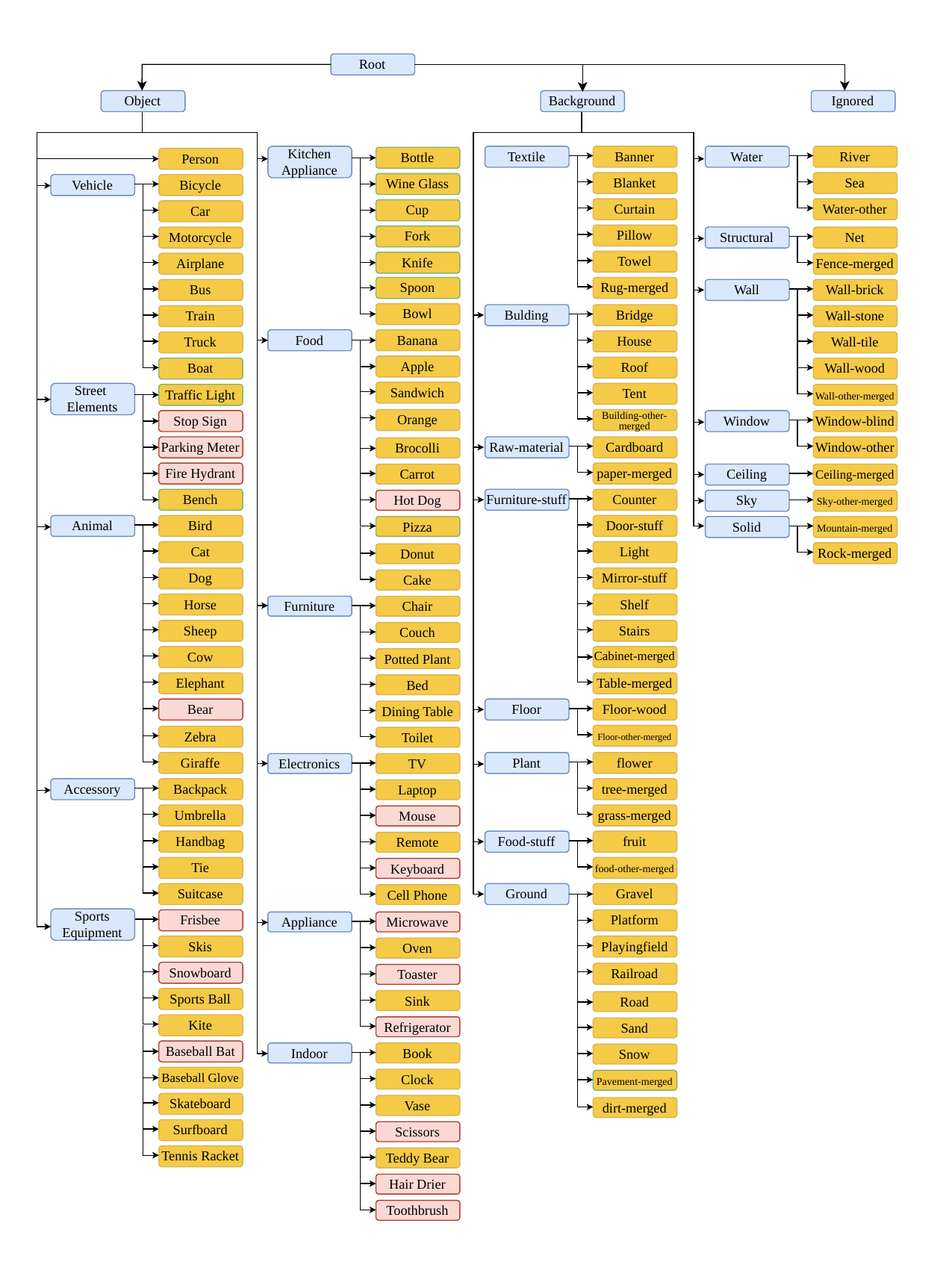}
\caption{Hierarchy definition for MS COCO~\cite{lin2014coco} with 4 levels. The classes that are considered unknown are marked in red. }
\label{fig:coco}
\end{figure}

Following~\cite{hwang2021eopsn, gasperini2023u3hs, schmidt2025prior2former}, the least frequent 20\% of \emph{thing} classes in MS COCO~\cite{lin2014coco} are treated as unknown. These classes are never observed during training, as images containing them are removed from the training set. The resulting training set contains 98,112 images with 117 remaining classes. The unknown classes are:
\emph{fire hydrant, stop sign, parking meter, bear, frisbee, snowboard, baseball bat, hot dog, mouse, keyboard, microwave, toaster, refrigerator, scissors, hair dryer, tooth brush}. For defining the hierarchy for MS COCO~\cite{lin2014coco}, we primarily follow the supercategories provided in the dataset (Figure~\ref{fig:coco}). Compared to Cityscapes, COCO contains substantially more classes and significantly more training images, which helps in learning more hierarchical, representative proxy embeddings.

\section{Additional Implementation Details}
In this section, we provide additional implementation details of our method.

\subsection{Hyperbolic Hierarchical Proxy Anchor Loss}

Our proposed hierarchy-aware loss builds upon the formulation of proxy anchor loss~\cite{kim2020proxy}. It is particularly suitable for our setting for two main reasons. First, its hardness-aware formulation emphasizes difficult positive and negative pairs through exponential weighting. Second, it provides explicit learnable proxy embeddings for each class, which serve as semantic anchors and later enable unknown semantic mining based on distances between query embeddings and proxy representations. Proxy anchor loss is originally formulated at the \emph{batch level}. Let $\mathcal{P}$ denote the set of all proxy embeddings, each corresponding to a ground-truth class. For an incoming mini-batch, the positive proxy set $\mathcal{P}^{+}$ contains all proxies corresponding to ground-truth classes appearing in the batch. The loss aggregates interactions between embeddings and proxies across the batch and is defined as
\begin{equation}
\begin{aligned}
\mathcal{L} =
&\frac{1}{|\mathcal{P}^{+}|}
\sum_{p \in \mathcal{P}^{+}}
\log \Big(1 + \sum_{x \in X_{p}^{+}} 
\exp(-\alpha (s(x,p) - \delta)) \Big) \\
&+
\frac{1}{|\mathcal{P}|}
\sum_{p \in \mathcal{P}}
\log \Big(1 + \sum_{x \in X_{p}^{-}} 
\exp(\alpha (s(x,p) + \delta)) \Big),
\end{aligned}
\end{equation}
where $X_{p}^{+}$ and $X_{p}^{-}$ denote the sets of positive and negative samples with respect to proxy $p$. However, the standard formulation assumes a flat class structure in which each sample is associated with only a single positive proxy. In our setting, classes are organized in a semantic hierarchy, where each node in the hierarchy tree is represented by a proxy embedding. Consequently, an embedding should be attracted not only to the proxy corresponding to its ground-truth class but also to the proxies associated with its ancestor concepts in the hierarchy. To explicitly model this hierarchical attraction, the hierarchy-aware loss in the main paper is expressed at the \emph{sample level}. For each sample, the positive and negative proxy sets are defined with respect to its ground-truth class. Specifically, the positive set consists of the proxy corresponding to the ground-truth class together with the proxies of all its ancestor concepts in the hierarchy, while the negative set contains all remaining proxies corresponding to unrelated concepts. In practice, we retain the batch-wise structure of proxy anchor loss in order to preserve its normalization scheme in which positive and negative contributions are normalized by the numbers of positive and negative proxies, respectively. For each incoming batch, we accumulate the positive contributions across all proxies that serve as positives and normalize them by the number of such proxies. Analogously, the accumulated negative contributions are normalized by the number of negative proxies. This preserves the batch-level normalization and hardness-aware weighting of proxy anchor loss while extending it to hierarchical supervision in hyperbolic space.

\subsection{Semantic Unknown Mining}

The structured hyperbolic embedding space allows us to estimate whether a query embedding $\mathbf{e}$, rejected by known classes, may still correspond to an unknown object. Since the hierarchy-aware loss organizes semantic concepts across different abstraction levels, hyperbolic distances can be used to infer the degree to which $\mathbf{e}$ aligns with object-consistent concepts. Our key assumption is that unknown objects may remain close to more abstract semantic concepts in the hierarchy. 

The hierarchy score $S_{\text{hier}}(\mathbf{e})$ captures the abstraction level of the semantic concept associated with $\mathbf{e}$. Let $\mathbf{p}^*(\mathbf{e})$ denote the nearest proxy of embedding $\mathbf{e}$ in hyperbolic space, and let $l$ denote its hierarchy level. For a hierarchy of depth four (H4), we denote the leaf nodes as level $l_0$, followed by progressively higher abstraction levels $l_1$, $l_2$, and the root $l_3$. The hierarchy score is defined as
\begin{equation}
S_{\text{hier}}(\mathbf{e}) = \sum_{l} \alpha_l \, \mathds{1}_l(\mathbf{e}),
\end{equation}
where $\mathds{1}_l(\mathbf{e})$ is the indicator function and equals $1$ if $\mathbf{p}^*(\mathbf{e})$ belongs to level $l$, and $0$ otherwise. In practice, we set $\alpha=(2,4,2)$ for hierarchy levels $l=\{0,1,2\}$, excluding the root level $l_3$ since it does not provide discriminative separation. In the H4 hierarchy, proxies at level $l_1$ provide the most informative semantic abstraction without losing object-level specificity. 

The final objectness score balances hierarchy-level abstraction and semantic divergence, where $\beta$ controls the contribution of the hierarchy term. We set $\beta=4$ for COCO and $\beta=1$ for Cityscapes, as the larger number of classes in COCO yields more informative ancestor representations.

To estimate how confidently an embedding aligns with its nearest semantic concept, we compute a margin-based divergence score. Specifically, we measure the distance gap between the closest proxy and competing proxies within different semantic sets. Let $\mathcal{P}_i$ denote a set of proxies (e.g., the object subtree or proxies at a specific hierarchy level). For each set, we compute a normalized margin
\begin{equation}
D_i(\mathbf{e}) =
\frac{d^{(i)}_{2}(\mathbf{e}) - d^{(i)}_{1}(\mathbf{e})}
{\sigma_i + \epsilon},
\end{equation}
where $d^{(i)}_{1}$ and $d^{(i)}_{2}$ denote the smallest and second-smallest distances between $\mathbf{e}$ and proxies in $\mathcal{P}_i$, and $\sigma_i$ denotes the corresponding distance variance used for normalization. The small constant $\epsilon$ prevents numerical instability when the variance becomes very small.

We compute four such divergences: one over the entire object subtree and three over proxy sets corresponding to hierarchy levels $l_0$, $l_1$, and $l_2$. The final divergence score aggregates these terms as
\begin{equation}
S_{\text{div}}(\mathbf{e}) = \sum_i w_i D_i(\mathbf{e}),
\end{equation}
where $\mathcal{P}_i$ corresponds to the object subtree and the proxy sets at hierarchy levels $l_0$, $l_1$, and $l_2$. In practice, we set $w=(2,3,3,0.5)$ for the full object subtree, and the hierarchy-level divergences as $l_0$, $l_1$, and $l_2$, respectively. The weights emphasize divergences at higher abstraction levels, reflecting that alignment with abstract semantic concepts is more informative for unknown object mining than alignment with fine-grained leaf categories.

\section{Additional Ablation Studies on Hyperbolic Curvature}

We further ablate the influence of the hyperbolic curvature parameter $c$. The curvature controls the geometry of the embedding space: larger values of $c$ correspond to a more strongly curved hyperbolic space, while $c=0$ reduces the formulation to the Euclidean case, where the hierarchical regularization degenerates to its Euclidean counterpart without hyperbolic computations.

As shown in Table~\ref{tab:curvature}, a moderate curvature of $c=0.1$ achieves the best performance on unknown classes while maintaining competitive open-world results. This choice is also common in hyperbolic vision works, where most computations are performed in Euclidean space, and embeddings are projected to hyperbolic space only in the final representation layer~\cite{ermolov2022hyperbolic,doan2024hypow, kim2023hier}. As the curvature increases further, the unknown detection performance drops while the performance on known classes slightly increases. In hyperbolic space, greater curvature causes distances to grow faster with increasing radius, leading to greater separation between embeddings. While this improves discrimination between known categories, it makes it harder to represent higher-level semantic ancestors near the root of the hierarchy, which are important for recognizing unknown objects. Finally, we compare FPS and peak GPU memory between the Euclidean setting ($c=0$) and the hyperbolic setting ($c=0.1$). Euclidean setting runs at 6.35 FPS with 6.30 GB peak memory, and hyperbolic setting runs at 6.51 FPS with 6.21 GB memory. This indicates that introducing hyperbolic geometry does not incur noticeable runtime and memory overhead.

\begin{table}[h]
\centering
\caption{
\textbf{Impact of hyperbolic curvature.} Models are trained on Cityscapes~\cite{cordts2016cityscapes} and evaluated on Lost\&Found~\cite{pinggera2016lost} (unknown classes) and Cityscapes (known classes) under the open-world setting. Moderate curvature ($c=0.1$) performs best on unseen classes, while larger curvature favors known-class separation. }
\setlength{\tabcolsep}{4pt}
\fontsize{10pt}{10pt}\selectfont

\begin{tabular}{c|c|ccc|ccc}
\toprule
\multirow{2}{*}{Manifold} &
\multirow{2}{*}{$c$} &
\multicolumn{3}{c|}{Unseen Classes} &
\multicolumn{3}{c}{Open-World} \\

& &
PQ$\uparrow$ & RQ$\uparrow$ & SQ$\uparrow$ &
PQ$\uparrow$ & RQ$\uparrow$ & SQ$\uparrow$ \\
\midrule

$\mathbb{R}$ & 0
& 7.17 & 11.13 & 69.36
& \textbf{56.28} & \textbf{67.80} & \textbf{81.67} \\

$\mathbb{H}$ & 0.1
& \textbf{11.33} & \textbf{16.06} & \textbf{70.55}
& 53.84 & 65.13 & 81.34 \\

$\mathbb{H}$ & 0.5
& 6.59 & 9.53 & 69.11
& 55.22 & 67.33 & 80.56 \\

$\mathbb{H}$ & 1
& 5.21 & 8.28 & 66.67
& 56.08 & 67.72 & 81.59 \\

\bottomrule
\end{tabular}
\label{tab:curvature}
\end{table}

\section{Additional Qualitative Results}

\subsection{Recognition of Unknown Objects}

\begin{figure}[h]
\centering
\includegraphics[width=\linewidth]{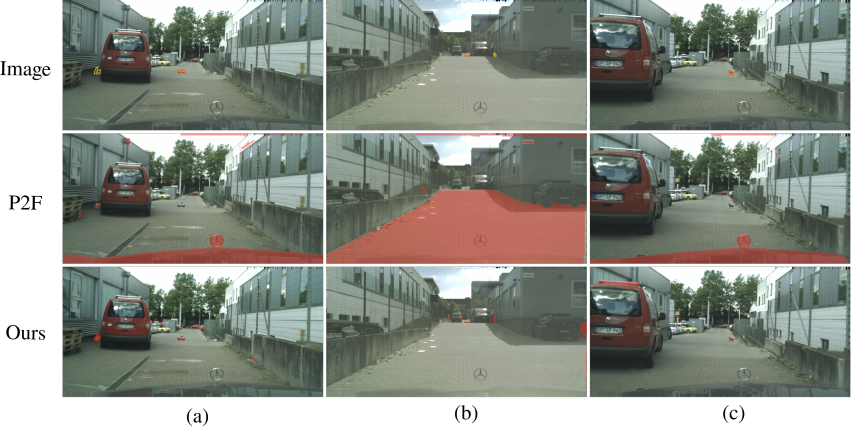}
\caption{\textbf{Qualitative comparison of unknown object prediction.} Images are from Lost\&Found~\cite{pinggera2016lost}. Annotated unknown objects are outlined in \textcolor{gt_unknown}{orange} and are reflected in the quantitative evaluation, while additional valid but unannotated ones are outlined in \textcolor{valid_unknown}{yellow}. Predicted unknown objects are shown in \textcolor{pred_unknown}{red}. Compared to P2F~\cite{schmidt2025prior2former}, which often merges OOD \textit{stuff} regions or misses unknown instances, our method produces more consistent instance-level predictions of unknown objects.}
\label{fig:unknown}
\end{figure}

We present additional qualitative comparisons between our method and P2F~\cite{schmidt2025prior2former} on detecting unknown objects in Figure~\ref{fig:unknown}. Compared to our approach, P2F often fails to detect certain unknown objects and occasionally misclassifies large \emph{stuff} regions as unknown instances. In contrast, our method produces more consistent predictions by leveraging the hierarchical embedding structure, which helps distinguish valid unknown objects from background regions.

\subsection{Open-set panoptic segmentation}

\begin{figure}[p]
\centering
\includegraphics[width=0.85\linewidth]{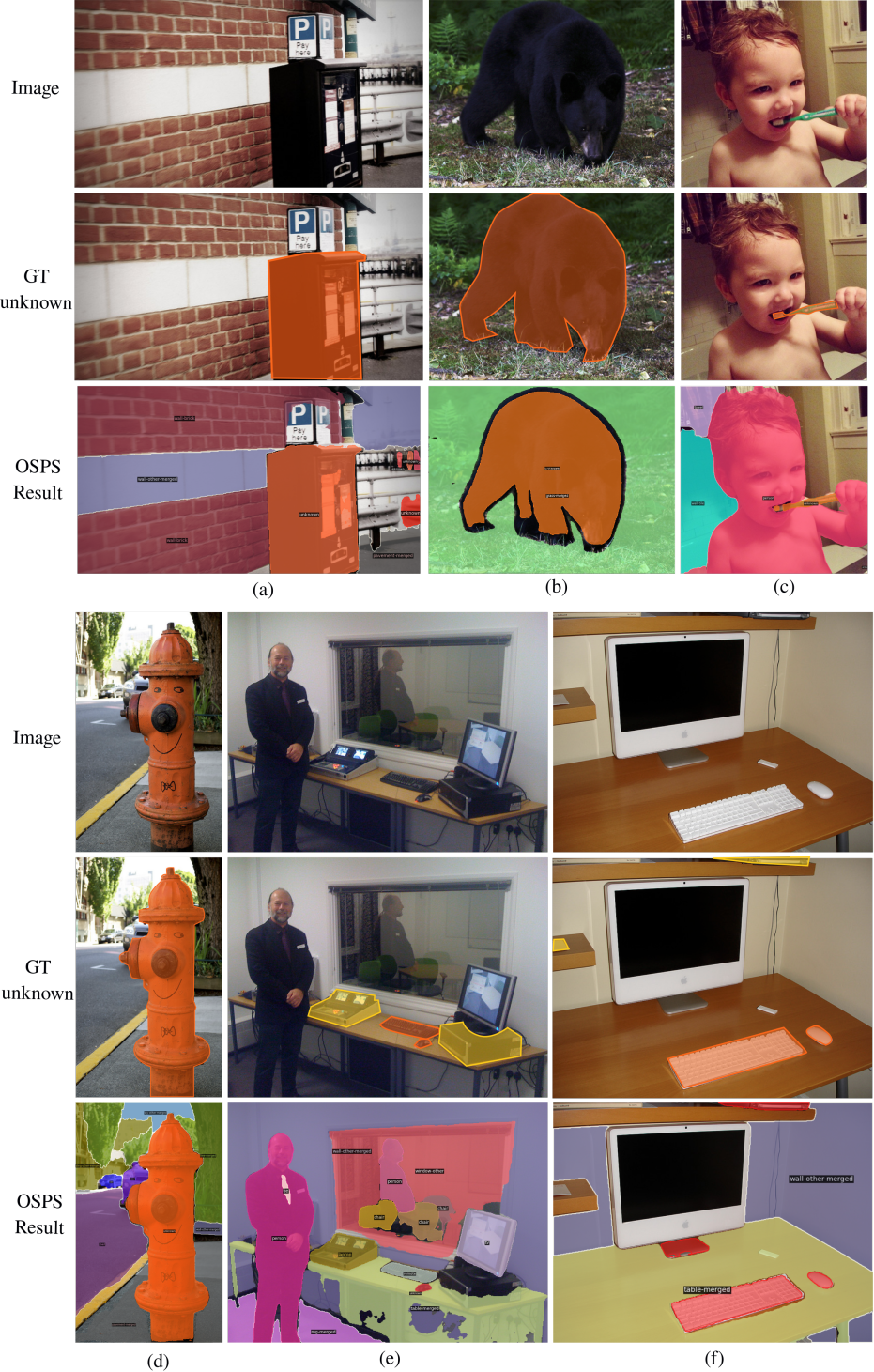}
\caption{\textbf{Open-set Panoptic Segmetation Results.} Images are from MS COCO~\cite{lin2014coco}. Annotated unknown objects are outlined in \textcolor{gt_unknown}{orange} and are reflected in the quantitative evaluation, while additional valid but unannotated ones are outlined in \textcolor{valid_unknown}{yellow}. Predicted unknown objects are shown in \textcolor{pred_unknown}{red}. Qualitative results demonstrating Hyp2Former’s ability to detect unknown objects while segmenting known classes.}
\label{fig:osps}
\end{figure}

In the following, we present additional qualitative results of Hyp2Former on open-set panoptic segmentation. As shown in Figure~\ref{fig:osps}, Hyp2Former is often able to generate accurate masks for unknown objects, as illustrated in examples Figure~\ref{fig:osps}(a)--(c). In more challenging cases, the model may occasionally misclassify unknown objects as known categories. For instance, the keyboard is mistaken for a remote and the control station is predicted as a laptop in Figure~\ref{fig:osps}(e). Nevertheless, Hyp2Former generally produces valid unknown object detections with well-defined masks while simultaneously segmenting known classes.

